\documentclass{article}

\usepackage{arxiv}

\usepackage[utf8]{inputenc} 
\usepackage[T1]{fontenc}    
\usepackage{url}            
\usepackage{booktabs}       
\usepackage{amsfonts}       
\usepackage{nicefrac}       
\usepackage{microtype}      
\usepackage{lipsum}
\usepackage{graphicx}
\usepackage[colorlinks=true,allcolors=blue]{hyperref}
\usepackage[super,comma,sort&compress]{natbib}
\usepackage{array}
\usepackage{longtable}
\usepackage{geometry}
\usepackage{endnotes}

\graphicspath{ {./images/} }

\title{GPT-LAB: next generation of optimal chemistry discovery by GPT driven robotic lab}

\author{
  Xiaokai Qin \\
  Research Center for Intelligent Sensing Systems\\
  Zhejiang Laboratory\\
    Hangzhou, Zhejiang 311121, China\\
  \texttt{qi0002ai@e.ntu.edu.sg} \\
   \And
   Mingda Song \\
   Research Center for Intelligent Sensing Systems\\
   Zhejiang Laboratory\\
     Hangzhou, Zhejiang 311121, China\\
   \texttt{u7386168@anu.edu.au} \\
  \And
  Yangguan Chen \\
  Research Center for Intelligent Sensing Systems\\
  Zhejiang Laboratory\\
    Hangzhou, Zhejiang 311121, China\\
  \texttt{chenyg@zhejianglab.com} \\
  \And
  Zhehong Ai \\
  Research Center for Intelligent Sensing Systems\\
  Zhejiang Laboratory\\
    Hangzhou, Zhejiang 311121, China\\
  \texttt{aizhehong20@mails.ucas.ac.cn} \\
  \And
  Jing Jiang\thanks{Corresponding author.}\\
  Research Center for Intelligent Sensing Systems\\
  Zhejiang Laboratory\\
    Hangzhou, Zhejiang 311121, China\\
  \texttt{jiangj@zhejianglab.com} \\
}

\begin{document}
\maketitle
\begin{abstract}
  The integration of robots in chemical experiments has enhanced experimental efficiency, but lacking the human intelligence to comprehend literature, they seldom provide assistance in experimental design. Therefore, achieving full-process autonomy from experiment design to validation in self-driven laboratories (SDL) remains a challenge. The introduction of Generative Pre-trained Transformers (GPT), particularly GPT-4, into robotic experimentation offers a solution. We introduce GPT-Lab, a paradigm that employs GPT models to give robots human-like intelligence. With our robotic experimentation platform, GPT-Lab mines literature for materials and methods and validates findings through high-throughput synthesis. As a demonstration, GPT-Lab analyzed 500 articles, identified 18 potential reagents, and successfully produced an accurate humidity colorimetric sensor with a root mean square error (RMSE) of 2.68\%. This showcases the rapid materials discovery and validation potential of our system.
\end{abstract}


\section{Introduction}
In recent years, the applications of self-driven laboratory (SDL) in fields of materials, chemical synthesis, biology, and medicine aroused widespread attention. SDLs typically includes the automation of experiments conduction with robotics and the usage of algorithm to guide robots to design, conduct and optimize the experiments. Leveraging the advantages of robotic executions in high consistency and accuracy, this application enables the generation of high-quality data for large scale search problems. The easier integration of advanced data-driven algorithms also surpassed human researchers in the context of high dimensionality and large scale of data processing. To some extent, this technique has showcased a significant acceleration in the discovery and design of new functional materials, synthetic paths, medicines \cite{king2004robot,zhang2020alloys,granda2018reactivity,wang2019copper,xue2016adaptive,wen2019design,raccuglia2016discovery,zhang2020phase}. However, most reported SDLs still relied on experienced scientist in designing and defining the search space for the robots in the first place, which includes searching and reviewing related literatures, propose possible product designs, and even some early stage manual experimental verifications. This still requires investing a considerable amount of time which can easily become the bottleneck for SDL applications \cite{burger2020robot,segler2018planning}.

Text mining holds the potential to expedite this process by preprocessing literatures for human experts. Many researchers have attempted to employ Natural Language Processing (NLP) techniques for text mining to extract information about experimental materials, protocols, and designs\cite{wang2022superalloy,peiz2023ultrahigh,zhou2018atoms,tshitoyan2019embeddings,nie2021graph,hakimi2020biomaterials,court2020phase,krenn2020research}. With the emergence of large language models (LLMs), researchers have explored their application in various traditional NLP tasks with promising results\cite{hoffmann2022models,lin2023protein,luo2022biogpt}. The advent of GPT-4, in particular, has garnered significant attention due to its impressive capabilities in language understanding and generation demonstrated across multiple domains\cite{openai2023gpt4}. Compared to traditional NLP models, GPT is better suited for literature mining and experimental protocol design with limited training data, demonstrating impressive performance even without extensive fine-tuning\cite{brown2020fewshot}.

Pioneering work in this direction has been conducted by researchers at CMU\cite{boiko2023research}, who showcased the application of GPT's Agent to assist scientific researchers in reading hardware documents and utilizing the Opentrons API\cite{opentronsAPI}. This experiment demonstrated the remarkable effectiveness of GPT-4 in the experimental design process, providing valuable insights into the potential of GPT for further exploration in the realm of experimental automation. Indeed, while there have been some promising initial attempts to utilize GPT for controlling experimental equipment and achieving success in that domain, fully replacing the role of researchers in collecting and referencing extensive literature for discovering novel reagents or materials has not been accomplished yet. The potential of GPT for assisting with literature text mining and material discovery is undeniable, given its language understanding capabilities and context-awareness. However, certain challenges remain to be addressed before GPT can autonomously and reliably identify new experimental reagents with minimal human intervention.

In this work, we designed a GPT enhanced SDL pipeline called ARMFE (Analysis - Retrieval - Mining - Feedback - Execution) to support automated end-to-end new material R\&Ds. An GPT-4-based agent was designed to empower the SDLs with the ability to rapidly and accurately explore research and development process experimental protocols based on the requests of human researchers and execute these protocols. As a proof of concept, we applied this pipeline on our self-built algorithm-guided robotic autonomous platform\cite{chen2023CO2} to develop new relative humidity (RH) colorimetric sensors, which rely on detecting color change of sensing materials upon exposure to target analyte. The agent demonstrated promising abilities to autonomously design experiment search space for the development of colorimetric sensing materials. A highly sensitive RH colorimetric sensor was developed and optimized via the pipeline and showed accurate prediction to RH with a root mean square error (RMSE) of 2.68\%. Our preliminary results have shown the potential of LLMs' application in empowering SDLs with robotic researchers capable of conducting independent R\&Ds with minimal human interventions.
\section{Method}
\begin{figure}[h]
  \centering
  \includegraphics[width=0.7\linewidth]{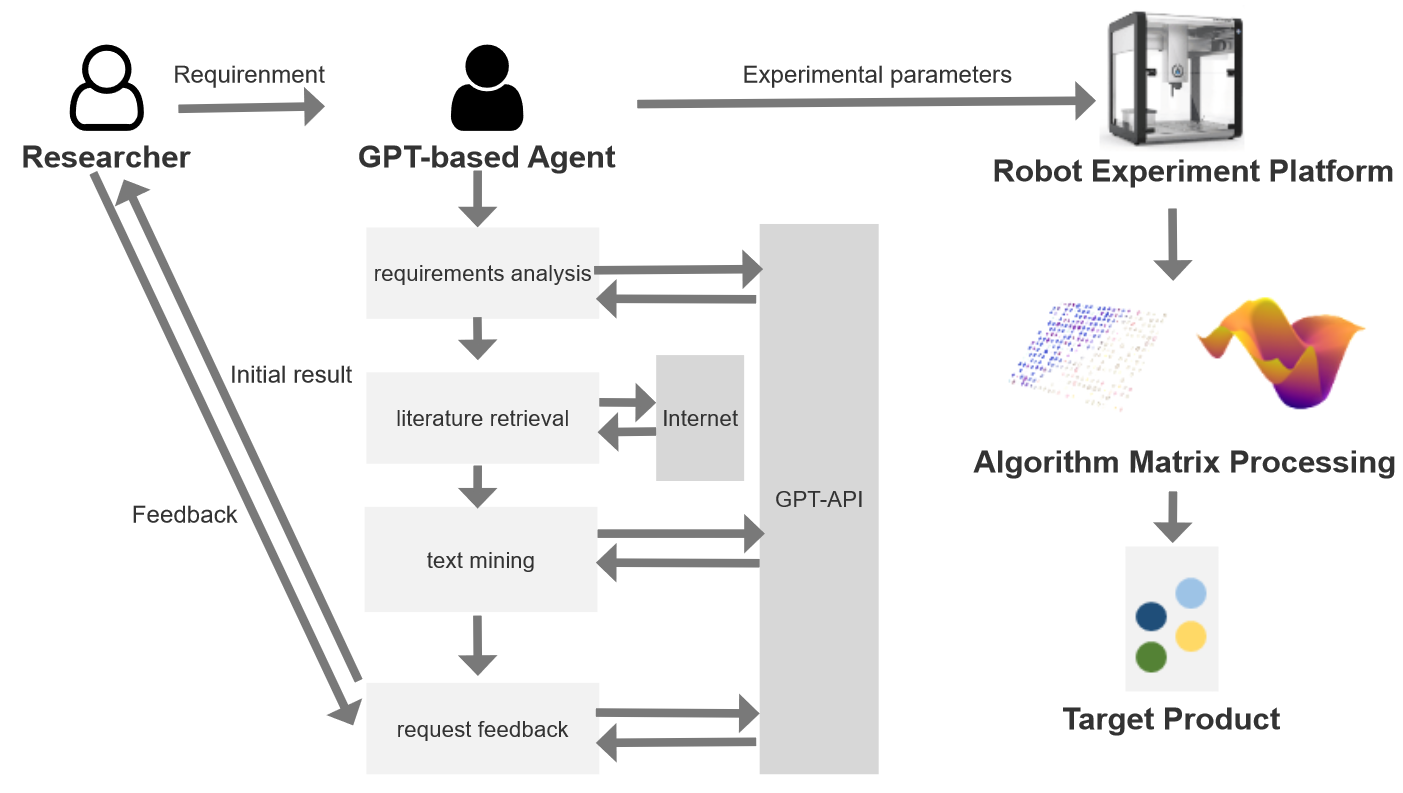}
  \caption{The ARMFE workflow of GPT-Lab}
  \label{fig:fig1}
\end{figure}

Our GPT-Lab is primarily comprised of two major modules: one is an automated experimental design agent based on the GPT framework, and the other is an algorithm-driven robotic experimentation platform. These two components synergistically establish a closed-loop automated workflow spanning from experimental requisites to empirical outcomes. The procedural workflow, which we refer to as ARMFE, is visually represented in Figure. \ref{fig:fig1}.The Agent follows a step-by-step process, including "requirements analysis," "literature retrieval," "text mining," "human researcher feedback," and "experiment execution," to establish a pathway from requirements to experimental results. Then the Agent conducted experiments using these substances.

\paragraph{Requirements analysis.}
In this phase, researchers are required to present a specific experimental requirement to the agent. We accomplish this task by utilizing the ChatGPT API. Through our meticulously crafted prompts and algorithms, the agent processes the provided experimental requisition to generate five pivotal keywords for conducting literature searches. In instances where the requisition lacks precision, the agent further engages with the researchers by posing clarifying questions, such as inquiring about the necessary methodologies to fulfill the experimental demands.
\paragraph{Literature retrieval.}
Upon obtaining the keywords derived from the initial step, the agent initiates an online search using these keywords to retrieve pertinent articles along with their respective titles and abstracts. Subsequently, the agent employs the ChatGPT API once again, utilizing the titles and abstracts of the articles to winnow out documents with relatively lower relevance. Following this filtration process, the agent procures complete articles from the internet to serve as the basis for subsequent analysis.
\paragraph{Text mining.}
We will employ the GPT to facilitate the comprehension of each individual article. Through this process, the agent will extract pertinent information regarding the utilized substances and their roles within the experiments, thereby constructing coherent textual passages. Subsequently, utilizing the GPT once more, the agent will analyze and organize these distilled textual segments into a JSON data structure. These foundational details will encompass: the substance's nomenclature, its associated CAS registry number, its functional significance within the experiment, and the role it assumes during the experimental procedure. These JSON files will be preserved for further processing in the subsequent phase.
\paragraph{Human researcher feedback.}
GPT-Lab will curate the experimental substance parameters in the form of highly comprehensible text from the JSON data structure, presenting this information to the researchers for their consideration. Following the selection process, the researchers will inform the agent regarding the chosen experimental substances. Subsequently, the agent, through a feedback loop, will generate experimental parameters structured in JSON format, which will be transmitted to the robotic experimentation platform for actual execution.

\paragraph{Experiment execution.}
The robotic platform will undertake the process of liquid formulation and subsequent iterative operations based on the experimental parameters furnished by the agent. The principles underlying this aspect will be expounded upon in greater detail in the forthcoming section.After the processing and designing of the GPT-based research agent, a material design space composing of the synthetic ingredients, their roles in the design, and typical compositions are proposed. For the sake of robotic platform execution, the Agent will compile a exchange file according the our robotic platform's requirements containing CAS codes for the experimental substances and their corresponding concentration values. This file will then be transmitted to the robotic experimentation platform for subsequent executions.

\section{Experiment}

\subsection{Article mining with Agent}
In evaluating Agent's efficiency in GPT-Lab, the system processed an average of 100 research articles within an hour. Utilizing multi-threading techniques, this processing speed can be augmented three to five-fold, representing a time-saving advantage of over a hundredfold compared to traditional manual extraction, especially when dealing with literature volumes in the tens of thousands. Furthermore, while human researchers may grapple with summarizing ultra-high dimensional variables, our system seamlessly integrates and analyzes a plethora of potential reagents, pinpointing those likely relevant to the research theme.

From the 500 articles analyzed, GPT-Lab identified fifty potential reagents. After filtration, 18 of these reagents with a relevance score of 80\% or higher were highlighted. Among them, there are 8 candidate core materials. For each, the system elucidated its experimental role, intended use, source, and rationale for its relevance. This curated list is then presented to researchers, allowing them to make informed selections based on their expertise and experimental needs.

Compared to pure GPT, our system has higher accuracy and better feasibility. Many substances provided by GPT cannot meet the conditions of subsequent robot experiments. In addition, GPT may hallucinate and generate output lacking theoretical basis. In contrast, the materials suggested by our GPT-Lab are more suitable and feasible for our experimental setup. At the same time, our system provides corresponding theoretical foundations and literature sources to back up its recommendations. The system automatically searches and obtains accurate material information focused on downstream robot experiments, rather than speculating without evidence. Compared to tedious manual literature searches, GPT-Lab greatly saves researchers time and effort while improving efficiency. Our system achieves full automation without needing human intervention at each step. Finally, after screening and selecting appropriate materials, the system seamlessly inputs them into downstream robots to complete the automated closed loop experimentation process.

\begin{figure}[h]
  \centering
  \includegraphics[width=0.7\linewidth]{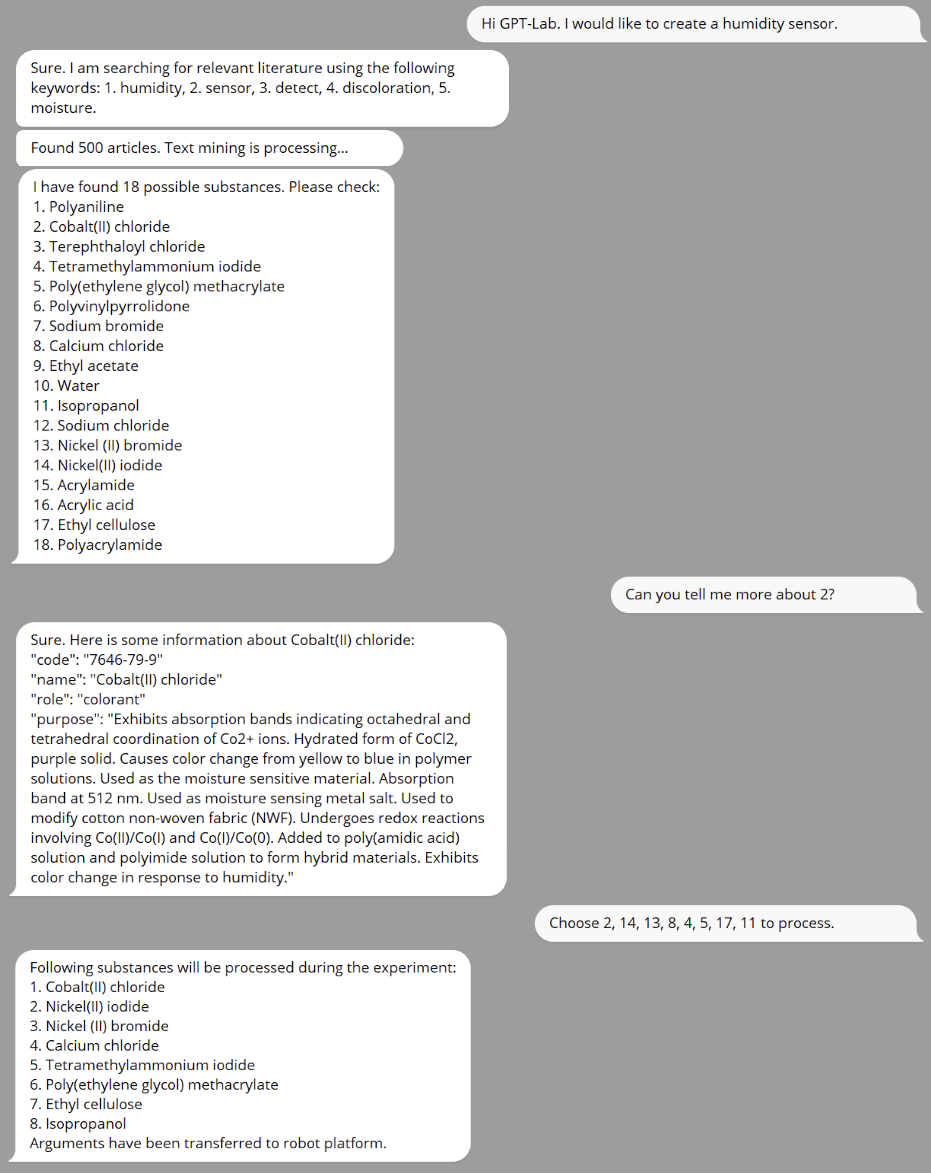}
  \caption{Dialogue demonstration with the agent}
  \label{fig:fig2}
\end{figure}

To demonstrate the versatility of our method, we explored applications beyond humidity sensor materials discovery. We utilized the system to search for key materials information for perovskite solar cells, obtaining a list of 7 materials including Methylammonium iodide, Lead iodide, Cesium Lead Chloride, Nickel oxide, Titanium dioxide, Lead bromide and Titanium tetrachloride. These materials serve purposes such as forming the perovskite structure, constituting the electronic band structure, acting as hole transport layer, and forming the TiO$_2$ layer. This again proves the ability of our approach for broad discovery and optimization of new materials.

Similarly, we explored a completely different problem - how to detect alkaloid content in mulberry leaves. We obtained some feasible key methods including using TSK gel Amide-80 column, C18 Phenomenex Synergi Fusion Stainless Steel Column for chromatographic separation, derivatizing the compound and separating it using different strategies, using MS to concurrently detect and identify the compound, and utilizing intra-day, inter-day, accuracy, precision, specificity and stability tests. After evaluation by domain experts, these methods were deemed viable and insightful for research in this field.

By achieving positive results across multiple distinct areas, we demonstrate that the method can be generalized for discovery of diverse materials and methods, not limited to a single application domain. This showcases the versatility of our approach and its potential to aid scientific discovery. We believe that with further research and refinement, this method can play an important role in a wide range of fields like materials science and drug development.

\subsection{Robotic experiment execution}
\begin{figure}[h]
  \centering
  \includegraphics[width=0.7\linewidth]{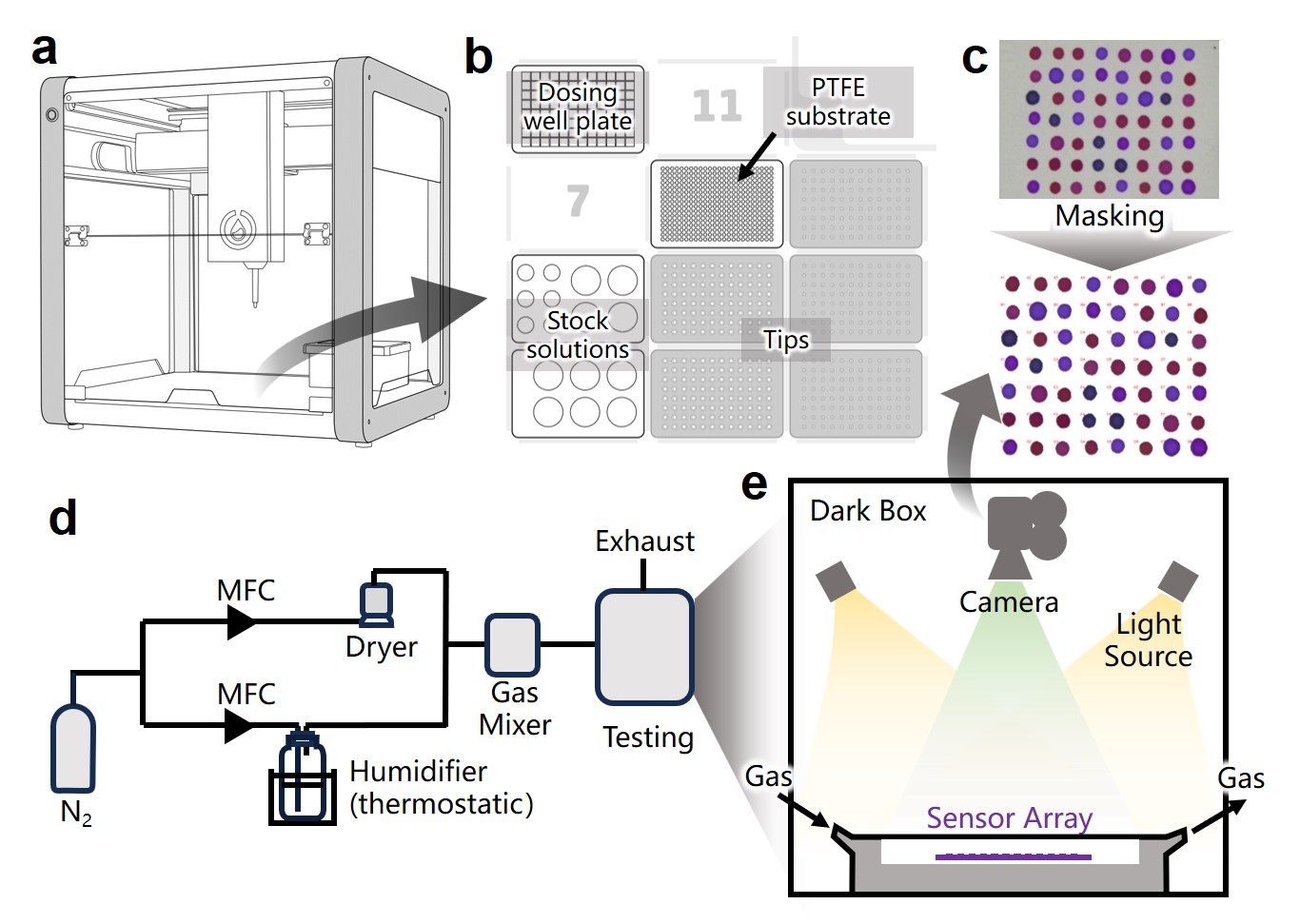}
  \caption{The robotic R\&D system. (a) Schematic of the liquid handling workstation. (b) The functional modules of the liquid handling workstation, including the stock solution area, pipette tip area, recipe configuration area, and sensing unit fabrication area. (c) Image of sensing units. Each color dot represents a gas-sensitive unit whose color is extracted by a computer vision algorithm. (d) Schematic of the gas path. The N$_2$ flow is split into two paths, passing through a dryer and a humidifier. The flow rates of the two streams are controlled by two MFCs and are mixed to achieve different RH for the gas-sensitive unit testing. (e) Gas testing setup, comprising a darkroom, light source, camera, and gas chamber. The gas-sensitive units are placed within the transparent upper chamber under uniform light conditions created by the darkroom and light source. The camera records the gas-sensitive units' color changes in different atmospheres.}
  \label{fig:fig3}
\end{figure}
The selected reagents are classified into three categories: colorants, additives, and solvents. Colorants include cobalt chloride (CoCl$_2$), nickel iodide (NiI$_2$), and nickel bromide (NiBr$_2$). Additives include calcium chloride (CaCl$_2$), tetramethylammonium iodide (TMAI), polyethylene glycol (PEG), and ethyl cellulose (EC). The solvent used is isopropanol (IPA). In the specific experiments, the dosage of each reagent is considered as a variable, resulting in a total of 8 variables. With a fixed total quantity, determining the dosages of the first 7 reagents automatically sets the dosage of the last reagent, resulting in a 7-dimensional variable space.

The execution of the wet experiment is close to the previously reported DBTM process\cite{chen2023CO2}.  The DBTM process is an efficient algorithm-guided process, implemented on a robot autonomous experiment platform, as shown in the Figure. \ref{fig:fig3}. According to user requirements, the optimal recipe can be obtained quickly by adjusting parameters. Specifically, the DBTM process involves a cycle of "recipe generation - robotic preparation - robotic testing - data processing - the next round of recipe generation". The robot equipment consists of a liquid handler and a self-built darkroom. Preparation steps are performed on the liquid handler, while testing is conducted in the darkroom. The testing involves flowing nitrogen (N$_2$) gas of different humidities to gas-sensitive samples fixed in the gas chamber. A camera continuously records their color changes under uniform lighting conditions, generating a color vs. time curve. From this curve, color change amplitude, response time, reversibility, sensitivity, and other indicators can be calculated. These indicators are weighted to derive the final score. The iterative process is guided by a Bayesian optimization algorithm, which determines the next round's sampling with higher uncertainty or higher potential of score improvement based on the inclination towards exploration or exploitation. In the specific experimental execution, 96 samples are collected in a single batch. While the first round's 96 recipes are randomly generated, the recipes for subsequent rounds are generated using Bayesian strategies. Each round includes a certain degree of exploration and exploitation tendencies.

\begin{figure}[h]
  \centering
  \includegraphics[width=0.7\linewidth]{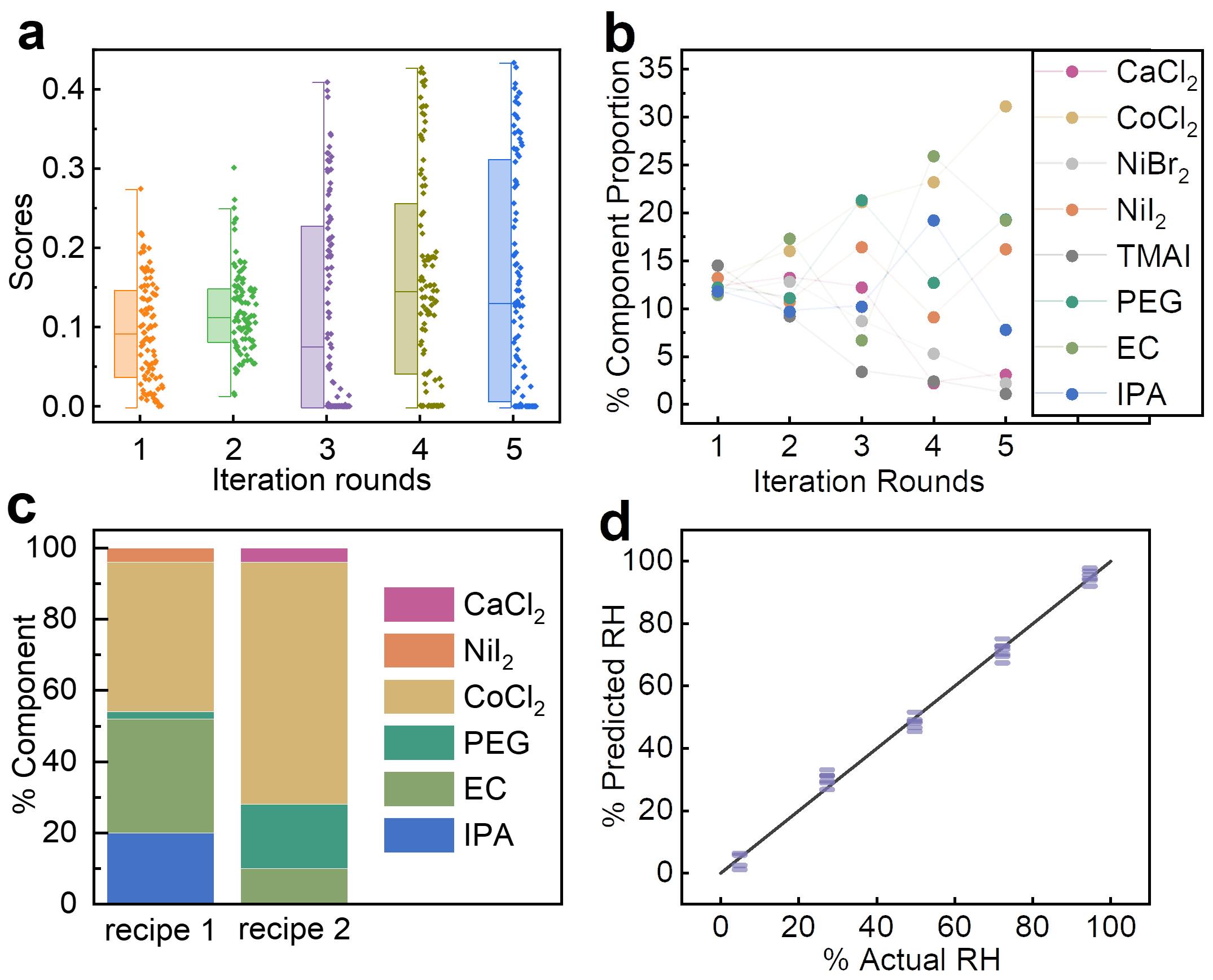}
  \caption{The wet experimental results. (a) The score distribution of recipes at different iteration rounds. (b) Total composition proportions of the 8 variables of 96 recipes in each batch. (c) Components of the selected two recipes. (d) The RH predicting accuracy of the array composed of the two selected recipes.}
  \label{fig:fig4}
\end{figure}

The distribution of sample scores in different batches of experiments is depicted in Figure. \ref{fig:fig4}a. With increasing rounds, the maximum score of each round gradually increased. Starting from the third round, a substantial portion of samples clustered around the 0-score range. This phenomenon was due to the iterative strategy's deliberated inclination towards exploration to prevent getting trapped in local optima. However, some recipes influenced by this exploratory tendency exhibit higher uncertainty and tend to have more extreme values, resulting in lower scores. After 5 rounds of experiments and the accumulation of 480 samples, the increase in the maximum score became less significant. It was also worth noting that the score distribution in the fifth round is broader, as more samples achieved higher scores, and these scores were edging closer to 0 compared to the previous round. This implied that finding superior recipes within the realm of heightened uncertainty is challenging, and the current optimal recipe was nearing a quasi-global optimum.

Figure. \ref{fig:fig4}b presents the total usage of each substance in the 96 recipes for each of the five iterative rounds. In the first round, as the recipes were generated randomly, although the proportions of each substance might vary significantly within specific recipes, their total proportions remained similar. As the iterations progressed, the usage of CoCl$_2$ exhibited an overall increasing trend, indicating a tendency for its utilization. Conversely, the usage of CaCl$_2$, NiBr$_2$, and TMAI showed a general decrease, gradually being discarded. This tendency suggested that recipes with more CoCl$_2$ may yield better results, while CaCl$_2$, NiBr$_2$, and TMAI had limited or even adverse effects. The selected two recipes were presented in Figure. \ref{fig:fig4}c, where NiBr$_2$ and TMAI were excluded. Recipe 1 included a small amount of NiI$_2$, while Recipe 2 incorporated a small amount of CaCl$_2$, enhancing sensitivity to low and high humidity conditions. The predictive accuracy of the array composed of these two recipes towards RH, as shown in Figure. \ref{fig:fig4}d, achieved precise quantification across RH from 5\% to 95\% at room temperature, with a root mean square error (RMSE) of 2.68\%.

\section{Conclusion}
The current iteration of GPT-Lab has been primarily engaged in three key endeavors: Firstly, it has demonstrated the commendable performance of GPT in the domain of experimental design. Secondly, it has substantiated the viability of an automated closed-loop process encompassing the trajectory from experimental requisition to empirical outcomes. Thirdly, this innovation has empowered chemists lacking computer science expertise to seamlessly harness the robotic platform for high-throughput experimentation, thereby augmenting experimental efficiency. As a proof of concept, a colorimetric humidity sensor was built within a week with minor human interference, capable of predicting the relative humidity at room temperature in the range of 5-95\% with an RMSE of 2.68\%. 

However, during the experimentation process, we have encountered several challenges: (a) GPT's intelligence is still limited, often leading to inaccuracies in its outputs. When inappropriate responses occur, it necessitates programmatic checks and retries to ensure the agent's robustness. This significantly escalates the cost of GPT utilization. (b) Despite the present capabilities of GPT-Lab in economizing researchers' time by circumventing extensive literature review and experimentation, GPT's capacity to acquire domain-specific knowledge outside the literature it has been exposed to remains limited. This imposes the need for researchers to manually filter experimental parameters. Potential remedies involve training larger models endowed with substantial chemical knowledge to supplant the current GPT, or exploring approaches like knowledge graphs and fine-tuning on extensive datasets to enhance GPT's breadth of knowledge. As larger models continue to evolve, the landscape of chemical research is poised to become increasingly efficient and streamlined.

\bibliographystyle{unsrt}


\section*{Acknowledgement}
This work was supported by funding from the Center-initiated Research Project of Zhejiang Lab (Grant No. 113015-AL2202). Special thanks to Longhan Zhang from Zhejiang Lab, for his assistance in reviewing and editing the manuscript.

\section*{Author}
\textbf{Xiaokai Qin} has a B.Sc. degree in Computer Science from University of Liverpool and B.Sc. degree in Information and Computing Science from Xi'an Jiaotong Liverpool University. Presently, he is pursuing his Master's degree in Artificial Intelligence at Nanyang Technological University, Singapore. Before his Master’s study, he worked as a machine learning intern at Zhejiang Lab.

\textbf{Mingda Song} graduated from the Australian National University with a Master's degree in Computing, specializing in machine learning. He is currently an intern at Zhejiang Lab, working on natural language process research.

\textbf{Yangguan Chen} has a B.Sc in Material Chemistry from Sun Yat-Sen University and a Ph.D. in Polymer Chemistry and Physics from Sun Yat-Sen University. His current research includes gas-sensing materials and sensing array construction. He is currently working at the Zhejiang Laboratory in Hangzhou, China.

\textbf{Zhehong Ai} is a Ph.D. candidate at Hangzhou Institute for Advanced Study of the University of Chinese Academy of Sciences. He has a Bachelor of Engineering degree from the College of Electronic Science and Engineering of Jilin University. His research interests include sensor technology and autonomous R\&D, and his current work involves the automatic AI (artificial intelligence) platform of colorimetric sensor development.

\textbf{Jing Jiang} received his Ph.D. from the Department of Electrical and Computer Engineering at the University of Illinois at Urbana-Champaign (UIUC) under the supervision of Prof. G. Logan Liu. After graduation, he co-founded a startup that uses mobile robots to conduct synthetic biology research and worked as a lecturer at UIUC. His research includes the nanofabrication and development of biosensors with the robotic platform. He has won multiple international prizes, including Vodafone Wireless Innovation Project and Nokia Sensing XChallenge based on smartphone-based sensors. Now he is a researcher (principal investigator) at Zhejiang Lab, Hangzhou, China, focusing on developing biology, materials, and sensors based on intelligent computing and robotic automation systems.
\newpage
\section*{Appendix}
\begin{longtable}{|c|c|c|m{6cm}|}
  \caption{The chemicals generated by GPT-Lab Agent}                                                                                   \\
  \hline
  Code       & Name                               & Role     & \multicolumn{1}{c|}{Purpose}                                            \\
  \hline
  \endfirsthead
  \multicolumn{4}{c}%
  {{\bfseries \tablename\ \thetable{} -- continued from previous page}}                                                                \\
  \hline
  Code       & Name                               & Role     & \multicolumn{1}{c|}{Purpose}                                            \\
  \hline
  \endhead
  \hline \multicolumn{4}{|r|}{{Continued on next page}}                                                                                \\ \hline
  \endfoot
  \hline
  \endlastfoot
  25233-30-1 & Polyaniline                        & reactor  & Used as the sensing material for humidity detection.                    \\
  \hline
  7646-79-9  & Cobalt(II) chloride                & colorant & Exhibits absorption bands; causes color change in polymer solutions.    \\
  \hline
  100-20-9   & Terephthaloyl chloride             & reactor  & Used to crosslink gelatin for humidity-sensitive capsules.              \\
  \hline
  25852-37-3 & Poly(ethylene glycol) methacrylate & additive & Used as analytes and causes color change of CP1.                        \\
  \hline
  9003-39-8  & Polyvinylpyrrolidone               & additive & Polymer layer for PVPP-based humidity sensors.                          \\
  \hline
  7647-15-6  & Sodium Bromide                     & adjuster & Controls humidity inside test chamber.                                  \\
  \hline
  141-78-6   & Ethyl acetate                      & reactor  & Used as analytes and solvent for characterizing optical properties.     \\
  \hline
  7732-18-5  & Water                              & adjuster & Used for humidity response testing and as analyte for humidity sensing. \\
  \hline
  7647-14-5  & Sodium Chloride                    & adjuster & Used to control relative humidity for testing.                          \\
  \hline
  13462-88-9 & Nickel (II) bromide                & colorant & Sensing thin-film layer for the humidity sensor.                        \\
  \hline
  7718-54-9  & Nickel(II) iodide                  & colorant & Main moisture-sensitive material.                                       \\
  \hline
  79-06-1    & Acrylamide                         & reactor  & Used to prepare humidity-responsive polymer networks.                   \\
  \hline
  79-10-7    & Acrylic acid                       & reactor  & Used to prepare humidity-responsive polymer networks.                   \\
  \hline
  9003-05-8  & Polyacrylamide                     & additive & Used as organogel for humidity-responsive films.                        \\
  \hline
  10043-52-4 & Calcium chloride                   & colorant & Used to remove moisture.                                                \\
  \hline
  75-58-1    & Tetramethylammonium iodide         & additive & Mixed with Nickel(II) iodide to improve sensitivity.                    \\
  \hline
  25322-68-3 & Polyethylene glycol                & additive & Used as a surfactant.                                                   \\
  \hline
  9004-57-3  & Ethyl cellulose                    & additive & Used to adjust the morphology of the substrate.                         \\
\end{longtable}

\end{document}